\documentclass{article}
\pdfoutput=1
\usepackage[utf8]{inputenc}
\usepackage{url}
\usepackage{hyperref}  
\usepackage{cleveref}
\usepackage{xstring}
\IfSubStr{\pdftexbanner}{2009}{%
\usepackage{siunitx}
}{%
\usepackage[load-configurations = version-1]{siunitx}
}
\usepackage{spconf}
\usepackage{graphicx}
\usepackage{amssymb,amsmath,bm}
\usepackage{textcomp}
\usepackage{xcolor}
\usepackage[font={small},skip=2pt]{caption}
\usepackage{listings}
\usepackage[tracking=true]{microtype}

\newcommand{\crnn}{RETURNN}

\graphicspath{ {figures/} }

\colorlet{punct}{red!60!black}
\definecolor{background}{HTML}{EEEEEE}
\definecolor{delim}{RGB}{20,105,176}
\colorlet{numb}{magenta!60!black}

\lstdefinelanguage{json}{
    columns=flexible,
    numbers=left,
    basicstyle=\footnotesize,
    stepnumber=0,
    belowskip=1pt,
    keepspaces=true,
    numbersep=1pt,
    breaklines=false,
    extendedchars=true,
    showstringspaces=true,
    breaklines=true,
    frame=single,
    literate=
     *{0}{{{\color{numb}0}}}{1}
      {1}{{{\color{numb}1}}}{1}
      {2}{{{\color{numb}2}}}{1}
      {3}{{{\color{numb}3}}}{1}
      {4}{{{\color{numb}4}}}{1}
      {5}{{{\color{numb}5}}}{1}
      {6}{{{\color{numb}6}}}{1}
      {7}{{{\color{numb}7}}}{1}
      {8}{{{\color{numb}8}}}{1}
      {9}{{{\color{numb}9}}}{1}
      {:}{{{\color{punct}{:}}}}{1}
      {,}{{{\color{punct}{,}}}}{1}
      {\{}{{{\color{delim}{\{}}}}{1}
      {\}}{{{\color{delim}{\}}}}}{1}
      {[}{{{\color{delim}{[}}}}{1}
      {]}{{{\color{delim}{]}}}}{1},
}

\def\baselinestretch{0.87}

\makeatletter
\renewcommand{\section}{\@startsection
  {section}%
  {1}%
  {}%
  {-0.7\baselineskip}%
  {0.3\baselineskip}%
  {}}%

\renewcommand{\subsection}{\@startsection
  {subsection}%
  {2}%
  {}%
  {-0.7\baselineskip}%
  {0.3\baselineskip}%
  {}}%

\renewcommand{\subsubsection}{\@startsection
  {subsubsection}%
  {3}%
  {}%
  {-0.7\baselineskip}%
  {0.3\baselineskip}%
  {}}%

\makeatother

\title{\crnn{}: The RWTH Extensible Training framework for Universal Recurrent Neural Networks}
\name{{\em Patrick Doetsch, Albert Zeyer, Paul Voigtlaender, Ilia Kulikov,
Ralf Schlüter, Hermann Ney
}}
\address{Human Language Technology and Pattern Recognition,
 Computer Science Department, \\
 RWTH Aachen University, 52062 Aachen, Germany \\
  {\small \tt \{doetsch,zeyer,voigtlaender,kulikov,schlueter,ney\}@cs.rwth-aachen.de}
}

\begin{document}

\maketitle
\begin{abstract}
  In this work we release our extensible and easily configurable neural network training software.
  It provides a rich set of functional layers with a particular focus on efficient training of
  recurrent neural network topologies on multiple GPUs.
  The source of the software package is public and freely available for academic research purposes and
  can be used as a framework or as a standalone tool which supports a flexible configuration.
  The software allows to train state-of-the-art deep bidirectional long short-term memory (LSTM) models
  on both one dimensional data like speech or two dimensional data like handwritten text and was 
  used to develop successful submission systems in several evaluation campaigns.
\end{abstract}
\section{Introduction}
  Recurrent neural networks (RNNs) and in particular LSTMs \cite{Hochreiter97} now dominate most sequential learning tasks
  including automatic speech recognition (ASR) \cite{sak2014lstm,zeyer2016lstm},
  statistical machine translation (SMT) \cite{bahdanau2014neural},
  and image caption generation \cite{vinyals2015show}.
  The training of deep recurrent neural networks is considerably harder compared to pure feed-forward
  structures due to the accumulation of gradients over time.
  For a long time there were only very few implementations of the methods and topologies that are required for
  RNN training.
  This changed rapidly when solutions for automatic differentiation and symbolic representations were
  combined into powerful computing libraries \cite{autograd}.
  In the machine
  learning community the most prominent example during that time was
  Theano \cite{bastien2012theano,bergstra+al:2010-scipy},
  which provides an extensive Python package to compute derivatives using symbolic mathematical expressions.
  While most of these packages allow to comfortably design neural network architectures on a low level, they do not serve as
  ready-to-use solutions for large scale tasks. Instead, their primary focus is generality in order to allow for various
  system designs without introducing constraints due to performance or usability issues. There is also naturally not
  much interest in getting the best performance for a particular hardware setup, but instead to keep compatibility
  at a maximum.

  \crnn{} draws on Theano as an additional layer on top of the Theano library
  which aims to fill in the gap between research oriented software packages
  and application driven machine learning software like Caffe \cite{caffe}.
  Our software provides highly optimized LSTM kernels
  written in CUDA, as well as efficient training in a multi-GPU setup.
  We simplify the construction of new topologies using a
  JSON based network configuration file while also providing a way to extend the software by functional layers.
  The software comes with few dependencies and it is furthermore tightly integrated into the RASR software package
  that is also developed at our institute \cite{rybach2011:rasr,wiesler2014:rasr}. The aim of this paper is to provide 
  an overview of the most important aspects of the software.

  The paper is organized as follows:
  In \Cref{sec:related} we give an overview over the software \crnn{} is based on,
  as well as competing implementations
  that were used for tasks in ASR.
  \Cref{sec:usage} describes how to design a neural network training setup within \crnn{}.
  \Cref{sec:engine} gives an overview of the components of the tool.
  \Cref{sec:ext} then provides further information on how to extend \crnn{} through additional functional layers. Finally
  we demonstrate the efficiency of \crnn{} empirically by comparing it to TensorFlow and Torch.

\section{Related Work} \label{sec:related}

Theano \cite{bastien2012theano} is a Python based framework for
symbolic mathematical tensor expressions with support for automatic differentiation.
Expressions are modeled in a computational dependency graph which can further be augmented through
an automatic optimization procedure. The implementation of each graph node is abstract and can be defined
for various types of hardware like CPUs or GPUs.
These properties make Theano particularly useful for neural network training tasks.
By providing the required building blocks Theano
allows to define complex connectionist structures that are fully differentiable.
Keras \cite{chollet2015} is a high-level Theano based framework
for data-driven machine learning.
It is maybe the most similar software package to \crnn{}.
      Keras started as a pure Theano based framework but now it also supports
      TensorFlow as back-end with minimal restrictions.
Similar projects that are built on top of the Theano library
include Lasagne \cite{theano-lasagne} and Blocks \cite{van2015blocks}.

TensorFlow is the most recent open source machine learning package by Google
\cite{tensorflow-whitepaper}.
It is actively developed and comes already with many predefined solutions
such as LSTMs, end-to-end systems and others.
TensorFlow is similar to Theano as it also works with symbolic computation graphs
and automatic differentiation.

Torch \cite{Collobert:Torch} uses the Lua programming language and
consists of many flexible and modular components that were developed by the community.
In contrast to Theano, Torch does not use symbolic expressions and all calculations are done explicitly.

Other notable frameworks are C++ -based Caffe \cite{jia2014caffe}, Python-based Neon \cite{nervana-neon}
and Brainstorm \cite{schmidhuber-brainstorm}.
In \cite{bahrampour2015comparative} a comparison between
Caffe, Neon, Theano, and Torch was done.
Task specific software packages like RASR \cite{wiesler2014:rasr} or Kaldi \cite{Povey_ASRU2011_2011}
which are both for developing speech recognition systems,
contain modules to train and decode ASR systems, including neural networks.
While the EESEN package \cite{MiaoGM15} extends Kaldi by adding rudimentary support for LSTMs, \crnn{} extends 
RASR to support various recurrent neural networks architectures in ASR systems.

\section{General Usage} \label{sec:usage}

\crnn{} provides a fully functional training software, which includes user interaction,
a multi-batch trainer and the possibility to extract the network activations for further processing.
No other dependencies besides Theano are required and network topologies will always run on CPU or GPU.
During execution, \crnn{} writes useful information with configurable verbosity
to the standard output and a log file. Network activations can be forwarded into a HDF5 \cite{hdf5}
file or directly be passed to the RASR decoder as described in \Cref{sec:engine}. It is further
possible to execute \crnn{} in a daemon mode which allows to access model evaluation using web services.

\subsection{Configuration} \label{sec:configuration}

Network architectures are described using a JSON format. Each network hereby is a map from
layer identification names to layer descriptions. A layer description is simply a dictionary
containing a \textit{class} parameter which specifies the layer class, an optional list of
incoming layers, and layer specific parameters. When constructing the network, \crnn{} looks
for a layer with a specified loss
and then recursively instantiates all layers that are directly
or indirectly connected to it.
See \Cref{fig:json} for an example of a bidirectional LSTM network.

The remaining configuration parameters are provided as simple parameters.
A typical configuration file contains the task, the path or descriptor of the input data, a learning rate
together with suitable adjustment methods, and information on how batches should be crafted.
The configuration parameters can also be merged into the network JSON description file or even
provided fully in Python format, such that a single configuration file can be used.
The Python format further allows to define custom layer types and other functions in the configuration file
and to use them in the network.
We provide some demo setups and configuration files together with the release of the software.

\begin{figure}[tbp]\centering
\begin{lstlisting}[language=json]
"fw_0" : { "class":"rec", "n_out":300,"direction":1 },
"bw_0" : { "class":"rec", "n_out":300,"direction":-1 },
"fw_1" : { "class":"rec", "n_out":300,"direction":1,
           "from" : ["fw_0", "bw_0"]},
"bw_1" : { "class":"rec", "n_out":300,"direction":-1,
           "from" : ["fw_0", "bw_0"]},
"output" : { "class":"softmax","from" : ["fw_1", "bw_1"]}
\end{lstlisting}
  \caption{An example network specification JSON file that realizes a bidirectional LSTM-RNN with two layers containing 300 nodes in forward and backward direction correspondingly.}
    \label{fig:json}
  \end{figure}

 \subsection{Layers} \label{sec:layers}

 Layers are the fundamental building blocks of \crnn{}. Each layer is a named class which is
 callable in the JSON description file by specifying its constructor parameters. We already provide
 a rich set of feed-forward layers including convolutional operations \cite{icfhr},
 and support the most common activation functions. Convolutional layers hereby make use of the efficient kernel implementations within CuDNN.
 Network behavior can further be augmented by using functional
 components like sampling and windowing. Output layers with various loss functions are available, including the
 cross-entropy, the mean-squared-error and Connectionist Temporal Classification (CTC).

 The main focus however lies in the recurrent layers.
 Different cell implementations including
 (one- and two-dimensional) LSTM, gated recurrent units (GRU) \cite{gru},
 associative LSTM \cite{assoc}
 and many more variants
 are available.
 Recurrent layers can further be connected by passing over the final state from one RNN to another one,
 allowing for encoder / decoder topologies with attention as described in \cite{bahdanau2014neural}.
 A configurable attention
 mechanism is available to calculate expected inputs from encoder networks.
 Using a similar method, these recurrent
 layers further allow for basic language modeling.
An example can be seen in \Cref{fig:json}.
 Several layers can further be composed into sub-networks and then used as regular layers, which allows to model high order and
 circular dependencies between layers.

\section{Engine}
\label{sec:engine}

\begin{figure}[tbp]\centering
  \resizebox{.47\textwidth}{!}{\includegraphics{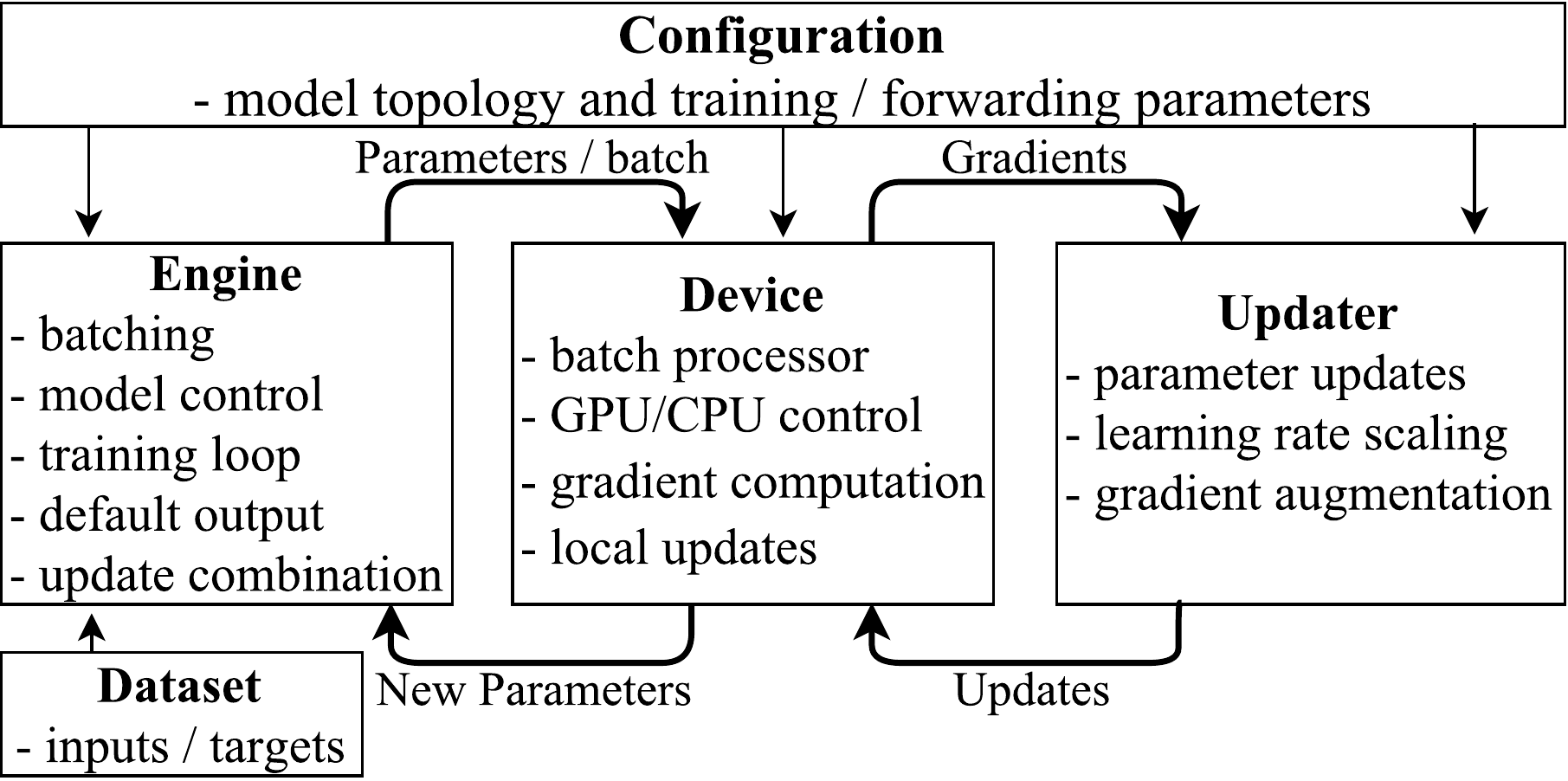}}
  \caption{The \crnn{} processing pipeline. Sequences are generated and passed
  over to the main engine. The engine combines sequences into batches and performs an epoch-wise
  training of the network parameters using one or more devices. Each device computes the
  error signal for its batch and generates updates according to the optimizer. }
   \label{fig:pipeline}
\end{figure}

Large vocabulary speech recognition tasks have memory
requirements that are significantly larger than the memory limitations of the
operating hardware.
This is particularly true for
GPUs. We therefore implemented a data caching technique that
minimizes hard disk usage while keeping the amount of allocated memory below a
configurable threshold. Also in many tasks the lengths of the sequences deviate by a large
amount and combining sequences into batches requires
to process many additional frames that were added by zero-padding. The software therefore
provides an option to chunk sequences into (possibly overlapping) segments of constant length.
By sacrificing contextual information provided in the end states of the chunks, chunking allows to
make a much more efficient use of the GPU memory \cite{doetsch:icfhr}. \crnn{} also supports a generic
pre-training scheme where simpler network topologies are automatically generated based on a given network topology.
A currently experimental Torch-Theano bridge which will be released with this software further
allows to run Torch code within \crnn{}.

In the recent release 0.8 of RASR \cite{rybach2011:rasr},
we added several generic Python interfaces, which allow to pass data in between
of RASR and \crnn{} efficiently. These interfaces can be used to perform
the feature extraction within RASR while passing the resulting inputs to the network
in real-time. They also provide a method to send the output activations of a network
to RASR in order to perform decoding or to retrieve an error signal which was calculated
based on discriminative training criteria available in RASR.

\subsection{Multi-GPU training}
\begin{figure}[tbp]\centering
  \resizebox{.3\textwidth}{!}{\includegraphics{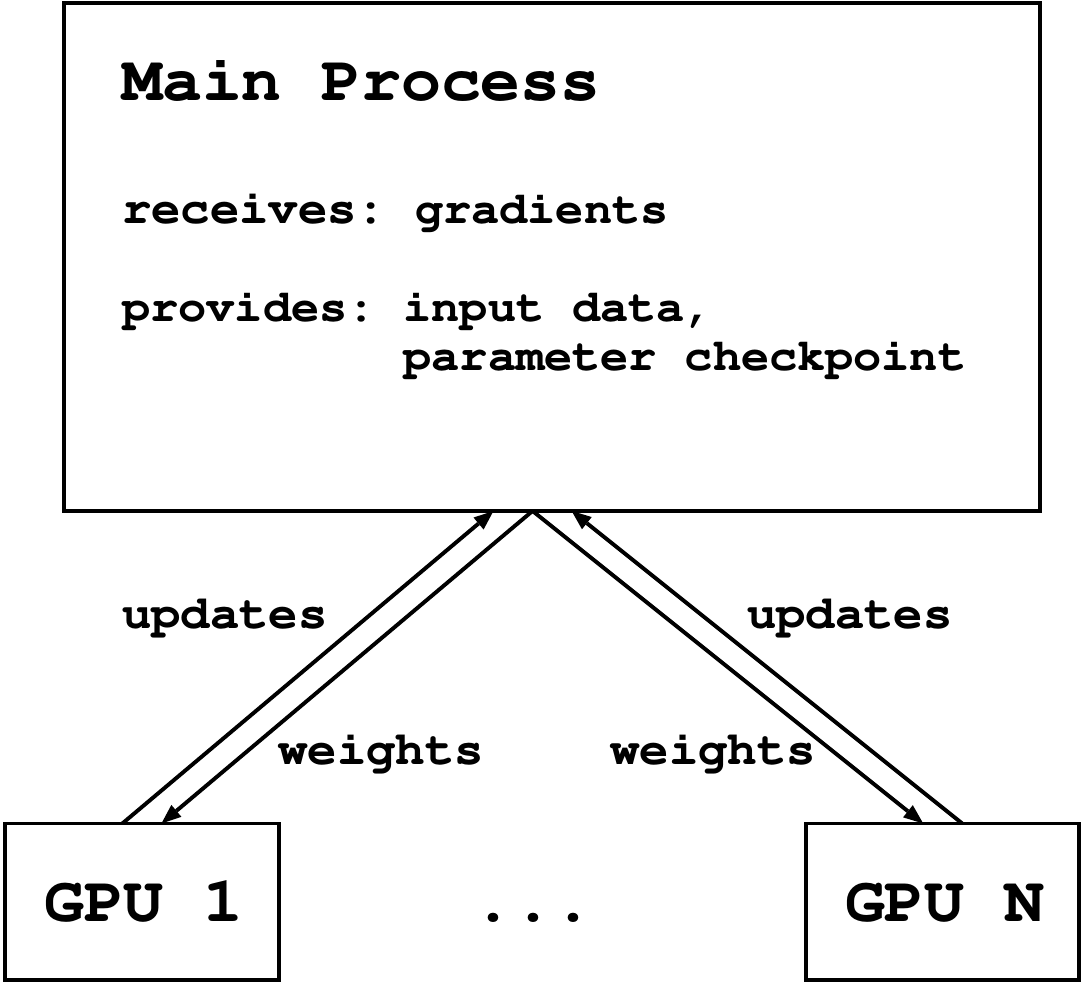}}
  \caption{The processing pipeline in multi-GPU training. An initial parameter set $\theta$ is passed to all
  workers. The workers make consecutive updates without synchronizing the parameters after every update. After processing
  three batches, the workers send their current parameter estimate back to the CPU process where they get combined into
  a single set of parameters.}
   \label{fig:mgpu}
\end{figure}

Modern machines consist of several GPUs where each of these
cards defines an isolated computation system. These computation systems can
be used as independent sub-batch processors.
Unfortunately, the library internally only allows to handle a single device context.
We therefore chose to implement the multi-GPU functionality as an interaction
of several independent system processes similar to \cite{NIPS2012_4687}.
Each GPU is attached to its own sub-process. Only the main process, which is scheduled
on the CPU, has access to the real network parameters. When data batches are processed,
a user specified number of batches is assigned to each device and the corresponding data
is copied to the GPU memory. The main process then provides an image of the current network
parameters to each of the GPU workers, which will apply their updates asynchronously batch by batch.
After processing a specific amount of batches the GPU workers send their modified network parameters
back to the main process where they are combined into a single set of parameters by averaging. The
overall process is depicted in \Cref{fig:mgpu}.
The processes communicate via sockets using a simple self-defined protocol. Weight matrices
are transferred as serialized arrays, which significantly slows down training if
the workers are synchronized too often. However, in our experiments we observe very stable
convergence even if we only synchronize once per epoch. In fact, we often observe a regularizing
effect and measure a better generalization error when keeping the GPUs asynchronous for several hundred
batches.

\subsection{CUDA Kernels for 1D and 2D LSTM Layers} \label{sec:cuda}

  We noticed that a straightforward LSTM implementation in Theano using scan (as used in Keras and Lasagne)
  is not very efficient in terms of both speed and memory. We therefore chose to implement the LSTM kernels directly
  using CUDA and cuBLAS \cite{icfhr}. The non-recurrent part of the LSTM forward computations are performed
  in a single matrix multiplication for the whole mini-batch of sequences. The same applies to the
  back propagation step with respect to the weights and the inputs, after the recurrent part is back propagated
  through time. Furthermore, we reuse memory wherever possible and use custom CUDA kernels for the LSTM gating
  mechanism. 

   To the best of our knowledge, we provide the first publicly available GPU-based implementation of multidimensional long short-term memory (MDLSTM) \cite{MDRNNs}. In an MDLSTM layer,
   the hidden state $h(u,v)$ for position $(u,v)$ is calculated based on the predecessor hidden activations $h(u-1,v)$ and $h(u,v-1)$, which means that it can only be computed after
   both predecessor states are known.
   As a consequence, previous (CPU-based) implementations of MDLSTM \cite{rnnlib} only process one pixel at a time by
   traversing the image column-wise in an outer loop and row-wise in an inner loop.
   We noticed that the activations for all positions on a common diagonal can be computed at the same time, which allows us to exploit the massive parallelism offered by modern GPUs.
   Additionally, we process multiple images and also the four directions of a multi-directional MDLSTM layer
   simultaneously by using batched cuBLAS operations and custom CUDA kernels. Optionally, the stable cell described in \cite{CellsLeifertSGL14} can be used to improve convergence.

\subsection{Optimization}
When optimizing deep RNNs, regular stochastic gradient descent may not allow the network to converge to a fixed point
in weight space and more sophisticated methods are needed. In those cases, learning rate scaling schedules aim to estimate a better
parameter dependent update step. In \crnn{} many well known learning rate schedules are implemented, including
Adagrad, Adadelta and Adam \cite{adagrad,adadelta,adam}. Furthermore \crnn{} allows for both the classical momentum term
and also the simplified Nesterov accelerated gradient \cite{momentum}. Decreasing the learning rate during training can be done based on the validation error.
In particular noise addition, norm constraints and outlier detection mechanisms allow for a better convergence and avoid numerical instabilities. Note that
batches gradients are not scaled in \crnn{} and the dimension of the batch has a direct influence on the norm of the gradients.
Regularization is possible using dropout on the layer inputs of any layer or by penalizing large L2 norms of the weight matrices.

\section{Extensibility} \label{sec:ext}

\crnn{} is mostly written in Python with some parts extended by modules using the C++ CUDA API
(see \Cref{sec:cuda}) and follows an object-oriented design. Any layer described in
\Cref{sec:layers} can be used as base class to extend the package by new functional elements.
Each layer is hereby considered as a black box that reads a batch of sequences and writes a batch of sequences, possibly of different shape.
In order to avoid influences of zero-padding when multiple sequences of different lengths are processed together, we use an index tensor
which indicates for each time step and batch, whether the frame should be considered as part of the sequence or not.
The layer definition itself can be any kind of Theano expression. Each layer
class is provided with a list of incoming layers and the index tensor.
The layer is expected to create a 3D tensor, with time (or sequence progress)
as first, the batch index as second and the layer output size as third dimension.
Likewise each layer in the list of incoming layers will provide a member called
\textit{output} with above defined shape.

A newly written layer class can directly be executed using the JSON description file, where
the variables of the corresponding JSON object are passed on as arguments to the constructor of the layer.

\subsection{Data Handling}
The dataset is abstracted as a generic interface.
Any dataset can provide multiple inputs and output targets
of variable dimensionality and shape, where inputs and outputs
can be encoded sparsely.
We have a wide range of dataset implementations.
Most prominently we support the HDF5 hierarchical data format, which is
also used as format for models produced in \crnn{}. Moreover, features
from RASR can directly be used within \crnn{} as described in \Cref{sec:engine}.
The release of this software contains several examples
of dataset usages.

\section{Experiments} \label{sec:experiments}
We demonstrate the performance of \crnn{} on frame-wise labeled speech data from the
CHiME dataset \cite{chime}.
Our aim is to show that \crnn{} successively converges during training and compare it to
other implementations. Each frame consists
of 17 consecutive speaker-adapted 16-dimensional MFCC vectors reduced to 45 dimensions by LDA. The vectors were labeled with 1501 tied allophone states using an Viterbi alignment obtained from a previously
trained hidden Markov model. The segments have an average length of 738 frames with a variance of 291 frames. To ease the processing in recurrent neural networks we divided all observation sequences into constant chunks of 250 frames, padding zero-frames if required. 
We compare \crnn{} to Keras, Torch and TensorFlow. In Torch we use a recently published CuDNN based LSTM implementation \footnote{https://github.com/soumith/cudnn.torch}, which is also planned to be migrated 
into Theano based frameworks.
For each package we measure the average runtime, average memory consumption and relative number of misclassified frames on 10\% of the training data. Training is performed on 81 chunks in parallel on a 
NVIDIA GTX 1080. Three bidirectional LSTM layers were used in the experiments containing 512 units in each direction. 
A similar configuration achieved a word error rate of 6.49\% and 8.43\% on the development set and evaluation set of the corpus.

  \begin{table}[tbp]
  \centering
  \caption{Comparison of runtime and memory requirement for different software packages. The numbers were averaged over 10 training epochs.
  Note that precise memory usage estimates for Torch and TensorFlow can not be obtained due to their internal memory management. }
  \label{tab:chime:compare}
  \begin{tabular}{lrrr}
    \hline
    \small Toolkit                 & \small Runtime [sec] & \small Memory [GB] & \small FER [\%] \\
    \hline
    \crnn{}                 &  198     & 2.4 &  42.51    \\
    \hline
    Theano LSTM             &  366     & 3.2 &  42.63   \\
    Keras                   &  619     & 5.4 &  44.36   \\
    TensorFlow              &  693     & $\sim$ 7.2 & 47.41    \\
    Torch (CuDNN)           &  164     & $\sim$ 2.6 & 43.02    \\
    \hline
  \end{tabular}
  \end{table}

It can be seen in \Cref{tab:chime:compare} that the internal LSTM kernel
of \crnn{} outperforms all competitors except the CuDNN implementation w.\,r.\,t.\,~runtime and memory usage. In order to provide a more direct comparison of the LSTM implementations, we also present the runtime of
\crnn{} with an LSTM version that does not make use of our optimized LSTM kernels (``Theano LSTM'' in \Cref{tab:chime:compare}). The internal memory management of TensorFlow and Torch make it difficult to obtain exact measurements
of their memory usage, but we can see that 25\% less memory is required in our LSTM kernel compared to the Theano based kernels, including Keras.

\begin{figure}[tbp]
  \centering
  \resizebox{.43\textwidth}{!}{\includegraphics{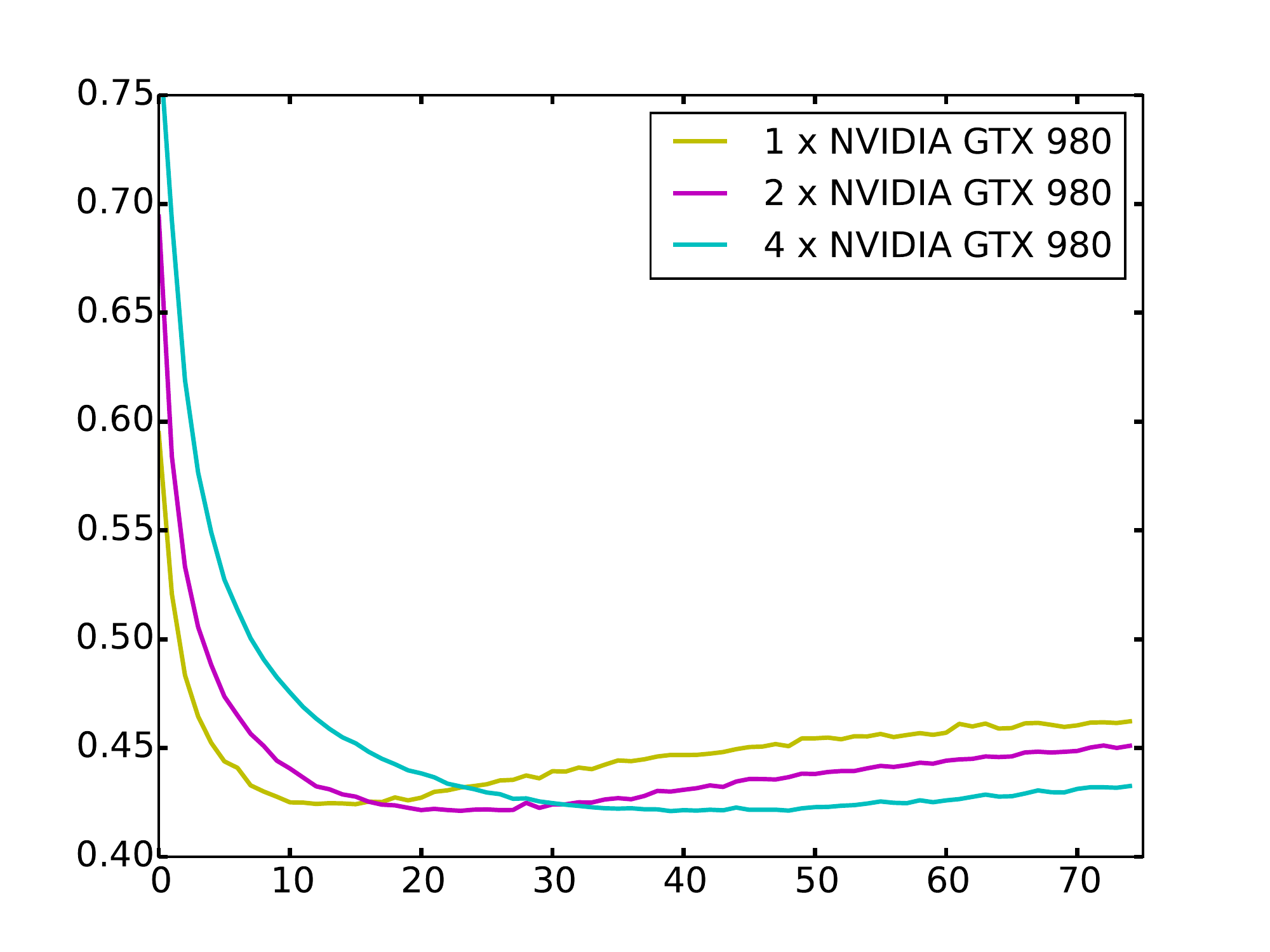}}
  \caption{Frame-error rate on the CHiME dataset over 75 epochs when training with multiple devices.
  The arrows indicate the minimal frame error and the total training time for one, two or four GPUs.}
  \label{fig:chime:mgpu}
 \end{figure}

We also conducted experiments to evaluate the runtime and classification performance of \crnn{} on multiple GPUs.
Here, the training time per epoch was 140, 80 and 41 seconds for
one, two or four NVIDIA GTX 980 GPUs respectively. The evolution of the frame error rate (FER) and the corresponding minima are shown in
\Cref{fig:chime:mgpu}. We can see that convergence time can be significantly
reduced by using multiple devices. We further observe a smoothing effect from the model averaging, such that the system trained on
four GPUs achieved the lowest frame error in this experiment.

\section{Conclusions}
  We presented \crnn{}, a highly configurable training framework for neural networks. The software is only based on Theano
  and CUDA and provides very fast training procedures for recurrent neural networks upon others. It further includes a rich set
  of functional layers which can be applied in new network designs using a convenient JSON syntax. 
  \crnn{} was successfully used in several recent evaluation campaigns, including the READ handwriting 
  competition on ICFHR 2016, the IWSLT 2016 German speech transcription task, and the CHiME speech separation and recognition
  challenge in 2016 where we ranked first, first, and second respectively.

  By providing an RNN training framework, which allows to train neural networks with minimal configuration effort, we hope to increase interest
  in this research area and to allow more people to access these methods. \crnn{} can be downloaded on our institute's website\footnote{https://www-i6.informatik.rwth-aachen.de/web/Software/index.html} and is freely
  available for academic research purposes.

 \section{Acknowledgements}

{
\let\normalsize\footnotesize\normalsize
\def\baselinestretch{0.5}

 This work was partially supported by the Intelligence Advanced Research Projects Activity (IARPA) via Department of Defense U.S. Army Research Laboratory (DoD/ARL) contract no.
 W911NF-12-C-0012. The U.S. Government is authorized to reproduce and distribute reprints for Governmental purposes notwithstanding any copyright annotation thereon.
 Disclaimer: The views and conclusions contained herein are those of the authors and should not be interpreted as necessarily representing the official policies or endorsements,
 either expressed or implied, of IARPA, DoD/ARL, or the U.S. Government. Additionally, the research was partially supported by Ford Motor Company and by the Deutsche Forschungsgemeinschaft
 (DFG) under contract no. Schl2043/11-1.
 We also would like to thank the authors of Theano \cite{bastien2012theano,bergstra+al:2010-scipy} for their tools enabling this work.
}

\vfill\pagebreak

\bibliographystyle{IEEEbib}

\def\baselinestretch{0.83}
\let\normalsize\footnotesize\normalsize

\let\OLDthebibliography\thebibliography
\renewcommand\thebibliography[1]{
  \OLDthebibliography{#1}
  \setlength{\parskip}{1pt}
  \setlength{\itemsep}{1pt plus 0.3ex}
}

\bibliography{strings,refs.bib}

\end{document}